\documentclass{article}

\usepackage[nonatbib,preprint]{neurips_2020}

\usepackage[utf8]{inputenc} 
\usepackage[T1]{fontenc}    
\usepackage{hyperref}     
\usepackage{url}            
\usepackage{booktabs}       
\usepackage{amsmath}
\usepackage{amsfonts}       
\usepackage{nicefrac}       
\usepackage{microtype}      
\usepackage{multirow}
\usepackage{caption}
\usepackage{subcaption}
\usepackage{ wasysym }
\usepackage{tikz}
\usepackage{algorithm}
\usepackage{wrapfig}
\usepackage{algorithmic}
\usepackage{amssymb}
\usepackage{pifont}
\usepackage{booktabs,arydshln}

\usetikzlibrary{arrows.meta}

\usetikzlibrary{calc}
\newtheorem{theorem}{Theorem}
\newtheorem{lemma}{Lemma}
\newtheorem{definition}{Definition}
\newcommand{\elmentwisemul}{\astrosun}%

\newcommand{\beginsupplement}{%
        \setcounter{table}{0}
        \renewcommand{\thetable}{S\arabic{table}}%
        \setcounter{algorithm}{0}
        \renewcommand{\thealgorithm}{S\arabic{algorithm}}%
        \setcounter{equation}{0}
        \renewcommand{\theequation}{S\arabic{equation}}%
        \setcounter{figure}{0}
        \renewcommand{\thefigure}{S\arabic{figure}}%
        \setcounter{section}{0}
        \renewcommand{\thesection}{S\arabic{section}}%
     }

\title{Learning Long-Term Dependencies in Irregularly-Sampled Time Series}

\newenvironment{itemize*}%
  {\begin{itemize}%
    \setlength{\itemsep}{0pt}%
    \setlength{\parskip}{0pt}}%
  {\end{itemize}}

\makeatletter
\def\adl@drawiv#1#2#3{%
    \hskip.5\tabcolsep
    \xleaders#3{#2.5\@tempdimb #1{1}#2.5\@tempdimb}%
    #2\z@ plus1fil minus1fil\relax
    \hskip.5\tabcolsep}
\newcommand{\cdashlinelr}[1]{%
    \noalign{\vskip\aboverulesep
        \global\let\@dashdrawstore\adl@draw
        \global\let\adl@draw\adl@drawiv}
    \cdashline{#1}
    \noalign{\global\let\adl@draw\@dashdrawstore
        \vskip\belowrulesep}}
\makeatother

\newlength{\commentindent}
\makeatletter
\renewcommand{\algorithmiccomment}[1]{\unskip\hfill\makebox[\commentindent][l]{$\vartriangleright$~#1}\par}
\LetLtxMacro{\oldalgorithmic}{\algorithmic}
\renewcommand{\algorithmic}[1][0]{%
  \oldalgorithmic[#1]%
  \renewcommand{\ALC@com}[1]{%
    \ifnum\pdfstrcmp{##1}{default}=0\else\algorithmiccomment{##1}\fi}%
}
\makeatother

\author{%
  Mathias Lechner \\
  IST Austria\\
  \texttt{mlechner@ist.ac.at} \\
  \And
  Ramin Hasani \\
  TU Wien \& MIT\\
  \texttt{rhasani@mit.edu} \\
}

\begin{document}

\maketitle

\begin{abstract}
Recurrent neural networks (RNNs) with continuous-time hidden states are a natural fit for modeling irregularly-sampled time series. These models, however, face difficulties when the input data possess long-term dependencies. We prove that similar to standard RNNs, the underlying reason for this issue is the vanishing or exploding of the gradient during training. This phenomenon is expressed by the ordinary differential equation (ODE) representation of the hidden state, regardless of the ODE solver's choice. We provide a solution by designing a new algorithm based on the \emph{long short-term memory} (LSTM) that separates its memory from its time-continuous state. This way, we encode a continuous-time dynamical flow within the RNN, allowing it to respond to inputs arriving at arbitrary time-lags while ensuring a constant error propagation through the memory path. We call these RNN models ODE-LSTMs. We experimentally show that ODE-LSTMs outperform advanced RNN-based counterparts on non-uniformly sampled data with long-term dependencies.
All code and data is available at \url{https://github.com/mlech26l/ode-lstms}.
\end{abstract}

\section{Introduction}
\begin{wrapfigure}[33]{r}{0.45\textwidth}
\vspace{0mm}
\centering
\includegraphics[width=0.45\textwidth]{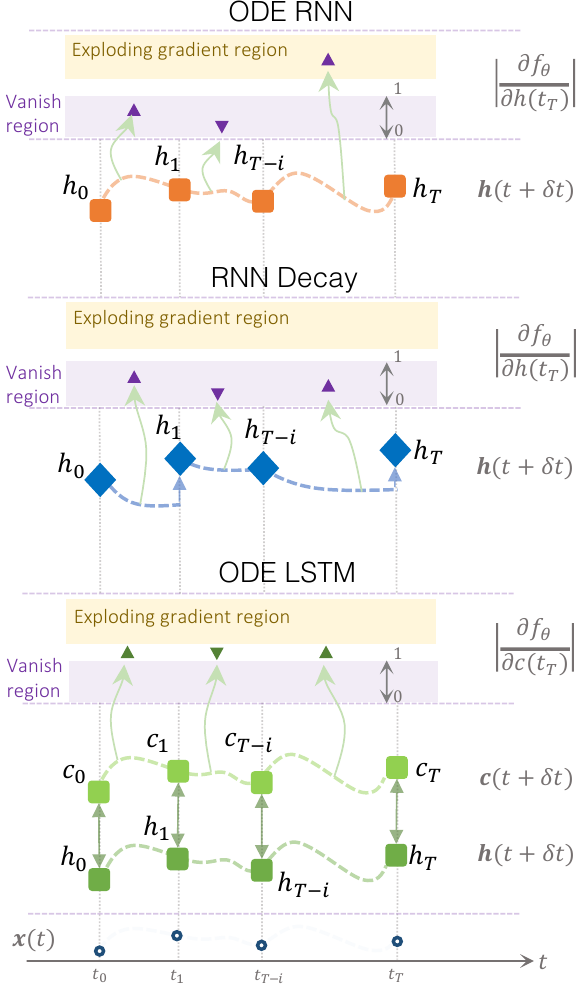}
\caption{Magnitude of the states' error propagation in time-continuous recurrent neural networks gives rise to the vanishing or exploding of the gradient (first two models). ODE-LSTMs are a solution to keep a constant gradient flow to avoid these phenomena in modeling irregularly sampled data.
}
\label{fig:the_figure}
\end{wrapfigure}

Irregularly-sampled time series, routine data streams in medical and business settings, can be modeled effectively by a time-continuous version of recurrent neural networks (RNNs). These class of RNNs whose hidden states are identified by ordinary differential equations, termed an ODE-RNN \cite{rubanova2019latent}, provably suffer from the vanishing and exploding gradient problem (see Figure \ref{fig:the_figure}, the first two models), when trained by reverse-mode automatic differentiation \cite{rumelhart1986learning,pontryagin2018mathematical}. 

An elegant solution to the vanishing gradient phenomenon \cite{hochreiter1991untersuchungen,bengio1994learning}, which results in difficulties in learning long-term dependencies in RNNs, is the long short term memory networks (LSTM) \cite{hochreiter1997long}. LSTMs enforce a constant error propagation through the hidden states, learn to forget, and disentangle the hidden states (memory) from their output states. Despite becoming the standard choice in modeling regularly-sampled temporal dynamics, LSTMs similar to other discretized RNN models, face difficulties when the time-gap between the observations are irregular.

In this paper, we propose a compromise to design a novel recurrent neural network algorithm that simultaneously enjoys the approximation capability of ODE-RNNs in modeling irregularly-sampled time series and capability of learning long-term dependencies of the LSTMs' computational graph. 

To perform this, we let an LSTM cell compute its implicit memory mechanism by their typical (input, forget, and output) gates while receiving their feedback inputs from a time-continuous output state representation. This way, we incorporate a continuous-time dynamical flow within the LSTM network, enabling cells to respond to data arriving at arbitrary time-lags, while avoiding the vanishing gradient problem, a model we call ODE-LSTMs (See Figure \ref{fig:the_figure}, the last model).

We compare ODE-LSTMs to standard and advanced continuous-time RNN variants, on a set of synthetic and real-world sparse time-series tasks, and discover consistently better performance. 

To put this in context, we first theoretically prove that the class of ODE-RNNs suffers from the exploding and vanishing gradient problem, making them unable to learn long-term dependencies efficiently. We show that learning ODE-RNNs by the adjoint method \cite{chen2018neural} does not help with this problem. As a solution, we propose ODE-LSTMs, a continuous-time RNN model capable of learning long-term dependencies of irregularly-sampled time-series.


\section{Background}

\textbf{ODE-RNNs   } Instead of explicitly defining a state update function, ODE-RNNs identify an ordinary differential equations in the following form \cite{funahashi1993approximation}:
\begin{equation}
\label{eq:ctrnn}
    \frac{\partial h}{\partial t} = f_{\theta}(x_{t+T},h_t,T) - \tau h,
\end{equation}
where $x_t$ is the input sequence, $h_t$ is an RNN's hidden state, and $\tau$ is a dampening factor. The time-lag $T$ specifies at what times the inputs $x_t$ have been sampled.

ODE-RNNs were recently rediscovered  \cite{rubanova2019latent} and have shown promise in approximating irregularly-sampled data, thanks to the implicit definition of \emph{time} in their resulting dynamical systems. ODE-RNNs can be trained by backpropagation through time (BPTT) \cite{rumelhart1986learning,werbos1988generalization,werbos1990backpropagation} through ODE solvers, or by treating the solver as a black-box and apply the adjoint method \cite{pontryagin2018mathematical} to gain memory efficiency \cite{chen2018neural}. In Section 3, we show this family of recurrent networks faces difficulty to learn long-term dependencies. 

\textbf{Long Short-term Memory   } LSTMs \cite{hochreiter1997long} express their discretized hidden states as a pair $(c_t,h_t)$ and its update function, $f_{\theta}(x_{t+1},(c_t,h_t),1) \mapsto (c_{t+1},h_{t+1})$ is defined as follows:

\begin{align}
    z_{t+1} &= \tanh(W_z x_{t+1} + R_z h_{t} + b_z) & \text{\textit{input update}}\label{eq:lstm1}\\
    i_{t+1} &= \sigma(W_i x_{t+1} + R_i h_{t} + b_i) & \text{\textit{input gate}}\\
    f_{t+1} &= \sigma(W_f x_{t+1} + R_f h_{t} + b_f + \mathbf{1}) & \text{\textit{forget gate}}\label{eq:forget}\\
    o_{t+1} &= \sigma(W_o x_{t+1} + R_o h_{t} + b_o) & \text{\textit{output gate}}\\
    c_{t+1} &= z_{t+1} \elmentwisemul i_{t+1} + c_{t}\elmentwisemul f_{t+1} & \text{\textit{cell update}}\\
    h_{t+1} &= \tanh(c_{t+1}) \elmentwisemul o_{t+1} & \text{\textit{output state}}\label{eq:lstm2},
\end{align}

where $\sigma$ is the sigmoid function $x\mapsto 1/(1+\exp(-x))$, the matrices $W_x$, $R_x$, and vectors $b_x$ for $x\in \{z,i,f,o\}$ are the weights of the RNN. The formulation shown in Equations (\ref{eq:lstm1}-\ref{eq:lstm2}) extends the original LSTM graph by a biased forget gate \cite{gers1999learning} (as implemented in PyTorch \cite{NEURIPS2019_9015} and TensorFlow \cite{tensorflow2015-whitepaper}). LSTMs demonstrate great performance on learning equidistant streams of data \cite{greff2016lstm}, however similar to other discrete-state RNNs, they are puzzled with the events arriving in-between observations. In Section 4, we introduce a continuous-time long short-term memory algorithm to tackle this.

\section{ODE-RNNs suffer from vanishing or exploding gradient}
In this section, we show that ODE-RNNs trained via backpropagation through time (BPTT) are susceptible to vanishing and exploding gradients. We also illustrate that the adjoint method is not immune to these gradient issues. We first formally define the gradient problems of the RNNs, and progressively construct Theorem \ref{thm:odevanish}.

\textbf{Gradient propagation in recurrent networks   }
Hochreiter \cite{hochreiter1991untersuchungen} discovered that the error-flow in the BPTT algorithm realizes a power series that determines the effectiveness of the learning process \cite{hochreiter1991untersuchungen,hochreiter1997long,bengio1994learning,pascanu2013difficulty}.
In particular, the state-previous state Jacobian of an RNN:
\begin{equation}\label{eq:jac}
    \frac{\partial {h_{t+T}(x_{t+T},h_t,T)}}{\partial h_t},
\end{equation}
governs whether the propagated error exponentially grows (explodes), exponentially vanishes, or stays constant. Formally:



\begin{definition}[Vanishing or exploding gradient]
    Let $h_{t+T} = f(x_{t+T},h_t,T)$ be a recurrent neural network, then we say unit $i$ of the network $f$ suffers from a vanishing gradient if for some small $\varepsilon>0$ it hold that
    \begin{equation}\label{eq:vanish}
    \Big| \sum_{j=1}^{N} \frac{\partial h^i_{t+T}}{\partial h^j_t}\Big| < 1-\varepsilon,
    \end{equation}
    where $N$ is the dimension of the hidden state $h_t$ and super-script $v^i$ denotes the $i$-th entry of the vector $v$.
    We say unit $i$ of the network $f$ suffers from an exploding gradient if it holds that
    \begin{equation}
    \Big| \sum_{j=1}^{N} \frac{\partial h^i_{t+T}}{\partial h^j_t}\Big| > 1.
    \end{equation}
    We say the whole network $f$ suffers from a vanishing or respectively exploding gradient problem if the above condition hold for some of its units.
\end{definition}

The factor $\varepsilon$ in Eq. \ref{eq:vanish} is essential as Gers et al. \cite{gers1999learning} observed that a learnable vanishing factor in the form of a forget-gate significantly benefits the learning capabilities of RNNs, i.e., the network can \emph{learn to forget}.
Note that a RNN can simultaneously suffer from a vanishing and an exploding gradient by the definition above. 

Now, consider an ODE-RNN given by Eq. \ref{eq:ctrnn} is implemented either by an Explicit Euler discretization or by a Runge-Kutta method \cite{runge1895numerische,dormand1980family}.  We can formulate their state-previous state Jacobian in the following two lemmas:

\begin{lemma}
\label{lem:1}
    Let $\dot{h} = f_\theta(x,h,T) - h \tau$ be an ODE-RNN. Then state-previous state Jacobian of the explicit Euler is given by the following equation:
    \begin{equation}
         \frac{\partial h_{t+T}}{\partial h_t} = I + T\frac{\partial f}{\partial h}\Big|_{h=h_t} - \tau T I.
    \end{equation}
\end{lemma}
\begin{lemma}
\label{lem:2}
    Let $\dot{h} = f_\theta(x,h,T) - h \tau$ be an ODE-RNN. Then state-previous state Jacobian of the Runge-Kutta method is given by
    \begin{equation}
          \frac{\partial h_{t+T}}{\partial h_t} = I + T\sum_{j=1}^{M} b_i \frac{\partial f}{\partial h}\Big|_{h=K_i} - \tau T I.,
    \end{equation}
    , where $\sum_{j=1}^{M} b_i=1$ and some $K_i$.
\end{lemma}

The proofs for Lemma \ref{lem:1} and Lemma \ref{lem:2} is provided in the supplements. Consequently, we have:

\begin{theorem}[ODE-RNNs suffer from a vanish or exploding gradient]\label{thm:odevanish}
    Let $\dot{h} = f_\theta(x,h,T) - h \tau$, and $h_t$ the RNN obtained by simulating the ODE by a solver based on the explicit Euler or Runge-Kutta method.
    Then the RNN suffers from a vanishing and exploding gradient problem, except for parameter configurations which give the non-trainable constant dynamics $f_\theta(h,x) = 0$, and cases where $f_\theta(h,x)$ is constant, for a particular input sequence $x$ and $\theta$.
\end{theorem}

The proof is given in full in the supplementary materials. A brief outline of the proof is as follow: 
First, we look a the special cases of $ \frac{\partial f}{\partial h}-\tau = 0$. While such $f$ would enforce a constant error propagation by making the Jacobians equal to the identity, it also removes all dynamics from the ODE state. In other words, it would operate the ODE as a memory element.
Intuitively, any interesting function $f_\theta$ pushes the Jacobians away from the identity matrix, creating a vanishing or exploding gradient depending on $f_\theta$.

\begin{theorem}
\label{thm:ode_solution}
\textbf{\emph{(ODE-RNNs suffer from a vanish or exploding gradient regardless of the choice of ODE-solver) }} 
    Let $\dot{h} = f_\theta(x,h,T) - h \tau$, with $f_\theta$ being uniformly Lipschitz continuous. 
    Moreover, let $h(t)$ be the solution of the initial value problem with initial state $h_0$.
    Then, the gradients $ \frac{\partial h(T)}{\partial h_0}$, i.e, the Jacobian of the ODE state at time $T$ with respect to the initial state $h_0$, can vanish and explode, except for parameter configurations which give rise to the non-trainable constant dynamics $f_\theta(h,x) = 0$, and cases where $f_\theta(h,x)$ constant, for a particular input sequence $x$ and parameters $\theta$.
\end{theorem}
The proof is given in full in the supplementary materials. A brief outline of the proof is as follow:
We start by approximating the initial-value problem by an explicit Euler method with a uniform step-size. 
We then let the step-size approach zero which due to the Picard–Lindelöf theorem, makes the series
converge to the true solution of the ODE.
Based on bounds on $\frac{\partial f}{\partial h}$, we can obtain bounds of the gradients in the limit, which can vanish or explode depending on $f_\theta$.

\textbf{Does the adjoint method solve the vanishing gradient problem?}
Adjoint sensitivity method \cite{pontryagin2018mathematical} allows for performing memory-efficient reverse-mode automatic differentiation for training neural networks with their hidden states defined by ODEs \cite{chen2018neural}. The method, however, possesses \emph{lossy} reverse-mode integration steps, as it forgets the computed steps during the forward-pass. Consequently, at each reverse-mode step, the backward gradient pass diverges from the true forward pass \cite{zhuang2020adaptive,gholami2019anode}. 
%
This is because the auxiliary differential equation in the adjoint sensitivity method, $a(t)$, still contains state-dependent components at each reverse-step, which depends on the historical values of the hidden states' gradient. Therefore, both vanilla BPTT and the adjoint method face difficulties for learning long-term dependencies. In the next section, we propose a solution.

\section{ODE-LSTM architecture}
The RNN state of a standard LSTM network, is represented by a pair $(c_t,h_t)$, where $c_t$ is the memory cell and $h_t$ the output state, i.e., see Equations (\ref{eq:lstm1}- \ref{eq:lstm2}). The memory $c_t$ ensures a constant error propagation and the output state $h_t$ enables the LSTM to learn non-linear dynamics. We modify the way the output state $h_t$ is computed while preserving its gating mechanisms and memory cell.

To perform this, we declare the output dynamics of a cell by a continuous-time representation, which realizes an ODE-RNN. This way, the output state depend on the elapsed time when processing irregularly sampled time-series. Nonetheless, as the LSTM gates receive feedback connections from the cells' outputs, the gating dynamics become dependent on the time-lag as well. The resulting architecture termed an ODE-LSTM is shown in algorithm \ref{algorithm:ode_lstm}. 

\begin{algorithm}[t]
\caption{The ODE-LSTM}
\label{algorithm:ode_lstm}
\begin{algorithmic}
\STATE \textbf{Input:} Datapoints and their timestamps $\{(x_t,t_i) \}_{i=1\dots N}$
\STATE \textbf{Parameters:} LSTM weights $\theta_l$, ODE-RNN weights $\theta$, output weight and bias $W_{output}, b_{output}$ 
\STATE $h_0 = \mathbf{0}$ \COMMENT{ODE state}
\STATE $c_0 = \mathbf{0}$ \COMMENT{Memory cell}
\FOR{$i = 1 \dots N$} 
\STATE $(c_i,h'_i) = \text{LSTM}(\theta_l,(c_{i-1},h_{i-1}),x_i)$
\STATE $h_i = \text{ODESolve}(f_\theta,h_{i-1},h'_i,t_t-t_{i-1}) $ \COMMENT{Post-process LSTM output by ODE-RNN}
\STATE $o_i = h_i W_{output} + b_{output}$
\ENDFOR
\RETURN $\{o_i\}_{i=1\dots N}$
\end{algorithmic}
\end{algorithm}

The fundamental distinction of ODE-LSTM to other variants is that they leave the RNN's memory mechanism untouched and assert the continuous dynamics into output function that processes the state. This way, ODE-LSTMs can learn long-term dependencies when trained by gradient descent. 

\begin{wraptable}[9]{r}{0.5\textwidth}
    \vspace{-4mm}
    \centering
    \caption{Change to the hidden states of an RNN between two observations $t$ and $t+T$}
    \vspace{-2mm}
    \begin{tabular}{lr}
        \toprule
        \textbf{Model} & State between observation \\
        \midrule
        Standard RNN & $h_t$ \\
        GRU-D & $h_t e^{-T \tau}$  \\
        ODE-RNN & ODE-Solve($f_\theta,h_t, T$)  \\
        ODE-LSTM & $\big(c_t, \text{ODE-Solve}(f_\theta,h_t, T)\big)$ \\
        \bottomrule
    \end{tabular}
    \vspace{1mm}
    \label{tab:states}
\end{wraptable}
On the contrary, recurrent network variants such as CT-RNN \cite{funahashi1993approximation}, continuous-time gated recurrent units (CT-GRU) \cite{mozer2017discrete}, and GRU-D \cite{che2018recurrent} incorporate the elapsed-time by a decay apparatus on the state, while preserving the rest of the RNN architecture. This decaying memory originates the vanishing factor during backward error-propagation, which results in difficulties in learning long-term dependencies. 

Our ODE-LSTMs are immune to this shortcoming.
More precisely, Table \ref{tab:states} lists how the transition of the hidden states between two observations of the ODE-LSTM differs from other architectures. 
Similar to the LSTM, we can ensure a near-constant error propagation at the beginning of the training process with a proper weight initialization.
\begin{theorem}
\label{thm:ode_lstm}
    Let $f$ with $(c_{t+T},h_{t+T}) = f(x_{t+T},(c_t,h_t),T)$ be an ODE-LSTM described by Algorithm 1. Moreover, we assume the weights  $R_z, R_i, R_f, W_f$ and $b_f$ are initialized close to 0.
    Then, the units $c_t$ of the state pair $(c_t,h_t)$ do not suffer from a vanishing or exploding gradient at the beginning of the training process.
    \vspace{-2.5mm}
\end{theorem}
The proof is given in full in the supplements. A brief outline: We assume that $R_z, R_i, R_f, W_f$ and $b_f$ are initialized close to 0 and we are at the beginning of the training, thus these values have not changed much yet. Consequently, we can neglect them and get $ \Big| \sum_{j=1}^{N} \frac{\partial {c^i_{t+T}(x_{t+T},(c_t,h_t),T)}}{\partial c^j_t} \Big|= \sigma(1) \approx 0.7310586$,  which is less than 1 (no exploding) but much greater than 0 (no vanishing). Note that exact value of the Jacobian at the beginning of the training can be controlled by the forget gate bias. If the underlying data express very long-term dependencies, we can increase the forget gate bias in Eq. (\ref{eq:forget}) and bring the error flow factor closer to 1. 

The ODE-LSTM can be viewed as a memory cell with gates controlled by a time-continuous process realized by ordinary differential equations. Next, we evaluate the performance of ODE-LSTMs in multiple time-series prediction tasks. 

\section{Experimental evaluation}
We constructed quantitative settings with synthetic and real-world benchmarks. We assessed the generalization performance of time-continuous RNN architectures on datasets that are deliberately created to express long-term dependencies and are of irregularly-sampled nature. All code and data is available at \url{https://github.com/mlech26l/ode-lstms}.

\begin{figure}[t]
    \centering
    \includegraphics[width=1\textwidth]{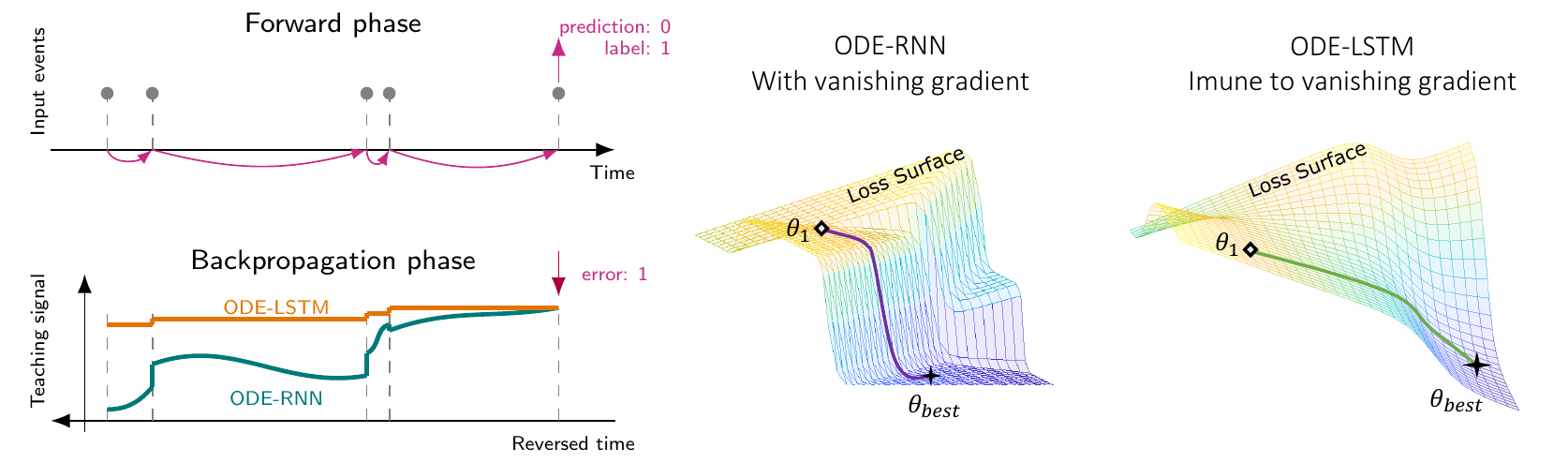}
    \caption{Left: Illustration of how vanishing gradients make the training process of RNNs difficult when the data express long-term dependencies. The prediction error can be thought of as a teaching signal indicating how the dynamics should be changed to minimize the loss. The vanishing gradient of the ODE-RNN makes the teaching signal weaker when propagating it back in time. Conversely, the teaching signal stays near-constant in the ODE-LSTM. Right: The resulting loss surfaces of the ODE-RNN is much flatter than ODE-LSTM, making the training difficult.}
\label{fig:loss_surface}
\end{figure}

\textbf{Baselines.} We compare ODE-LSTM to a large variety of continuous-time RNNs introduced to model irregularly-sampled data. This set includes RNNs with continuous-state dynamics such as ODE-RNN \cite{rubanova2019latent} and CT-RNNs \cite{funahashi1993approximation}, state-decay mechanisms such as CT-GRU \cite{mozer2017discrete}, RNN Decay \cite{rubanova2019latent}, CT-LSTM \cite{mei2017neural}, and GRU-D \cite{che2018recurrent}, in addition to oscillatory models such as Phased-LSTM \cite{neil2016phased}. 

Furthermore, we tested ODE-LSTMs against intuitive time-gap modeling approaches we built here, termed an augmented LSTM topology as well as bi-directional RNNs \cite{schuster1997bidirectional}. Experimental settings are given in the supplements.

\begin{table}[b]
    \centering
    \caption{\textbf{Bit-stream sequence classification}}
    \begin{tabular}{lcc}
    \toprule
        \multirow{2}{*}{Model} & Dense & Event-based  \\
                               & encoding  & encoding\\
        \hline
        ODE-RNN  & 50.47\% $\pm$ 0.06 & 51.21\% $\pm$ 0.37 \\
        CT-RNN & 50.42\% $\pm$ 0.12 & 50.79\% $\pm$ 0.34 \\
        Augmented LSTM  & \textbf{100.00\% $\pm$ 0.00} & 89.71\% $\pm$ 3.48 \\
        CT-GRU  & \textbf{100.00\% $\pm$ 0.00} & 61.36\% $\pm$ 4.87 \\
        RNN Decay  & 60.28\% $\pm$ 19.87 & 75.53\% $\pm$ 5.28 \\
        Bi-directional RNN   & \textbf{100.00\% $\pm$ 0.00} & 90.17\% $\pm$ 0.69 \\
        GRU-D   & \textbf{100.00\% $\pm$ 0.00} & {97.90}\% $\pm$ 1.71 \\
        PhasedLSTM   & 50.99\% $\pm$ 0.76 & 80.29\% $\pm$ 0.99 \\
        GRU-ODE & 50.41\% $\pm$ 0.40 & 52.52\% $\pm$ 0.35\\   
        CT-LSTM & 97.73\% $\pm$ 0.08 & 95.09\% $\pm$ 0.30\\
        \cdashlinelr{1-3}
        ODE-LSTM (ours)  & \textbf{100.00\% $\pm$ 0.00} & \textbf{98.89\% $\pm$ 0.26} \\
        \bottomrule
    \end{tabular}
    \caption*{\footnotesize \textbf{Note:} Test accuracy (mean $\pm$ std, $N=5$). While all of the above RNN can represent the correct function, training is difficult due to long-term dependencies.}
    \label{tab:synthetic}
\end{table}
\subsection{Synthetic benchmark - Bit-stream sequence classification}
We formulated a modified time-series variant of the XOR problem \cite{marvin1969perceptrons}. 
In particular, the model observes a block of binary data in the form of a bit-after-bit time-series. The objective is then to learn an XOR function of the incoming bit-stream. This setup is equivalent to the binary-classification of the input sequence, where the labels are obtained by applying an XOR function to the inputs. 



While any non-linear recurrent neural network architecture can learn the correct function, training the network to do so is non-trivial. For the model to make an accurate prediction, all bits in an upcoming chunk are required to be taken into account. However, the error signal is only provided after the last bit is observed. Consequently, during learning, the prediction error needs to be propagated to the first input time-step to precisely capture the dependencies, (see Figure \ref{fig:loss_surface}).

We designed two modes, a dense encoding mode in which the input sequence is represented as a regular, periodically sampled time-series, and an event-based mode which compresses the data into irregularly sampled bit-streams, e.g., $1,1,1,1$ is encoded as $(1,t=4)$. 
(See Table \ref{tab:synthetic}). We observed that a considerable number of RNN variants faced difficulties in modeling these tasks, even in the dense-encoding model. 

In particular, ODE-RNNs, CT-RNNs, RNN-Decay, Phased-LSTM, and GRU-ODE could not solve the XOR problem in the first mode. Phased-LSTM and RNN-Decay improved their performance in the second modality, whereas ODE-RNNs, CT-RNNs, and GRU-ODE still could not solve the task. The core reason for their mediocre performance is the exploitation of the vanishing gradient problem during training. The rest of the RNN variants (except CT-GRU) were successful in solving the task in both modes, with ODE-LSTM outperforming others in an event-based encoding scenario. 


\subsection{Person activity recognition with irregularly sampled time-series}

We consider the person activity recognition dataset from the UCI repository \cite{dua2019}. This task's objective is to classify the current activity of a person, from four inertial measurement sensors worn on the person's arms and feet. Even though the four sensors are measured at a fixed period of 211ms, the random phase-shifts between them creates an irregularly sampled time-series. Rubanova et al. \cite{rubanova2019latent} showed that ODE-based RNN architectures perform remarkably well on this dataset. Here, we benchmarked the performance of the ODE-LSTM model against other variants.

\begin{figure}[!t]
  \begin{minipage}{\textwidth}
  \vspace{0mm}
  \begin{minipage}[b]{0.45\textwidth}
    \centering
 \begin{table}[H]
    \centering
    \caption{\textbf{Per time-step classification}. Person activity recognition. Test accuracy (mean $\pm$ std, $N=5$)}
    \vspace{0mm}
    \begin{tabular}{lc}
    \toprule
    Model & Accuracy \\
    \hline
    ODE-RNN  & 80.43\% $\pm$ 1.55 \\
    CT-RNN  & 83.65\% $\pm$ 1.55 \\
    Augmented LSTM  & \textbf{84.11\% $\pm$ 0.68} \\
    CT-GRU & 79.48\% $\pm$ 2.12 \\
    RNN Decay & 62.89\% $\pm$ 3.87 \\
    Bi-directional RNN  &  \textbf{83.85\% $\pm$ 0.45} \\
    GRU-D  &  83.57\% $\pm$ 0.40 \\
    PhasedLSTM  &  83.33\% $\pm$ 0.69 \\
    GRU-ODE  & 82.56\% $\pm$ 2.63 \\
    CT-LSTM &  \textbf{84.13\% $\pm$ 0.11} \\
    \cdashlinelr{1-2}
    ODE-LSTM (ours)  &  \textbf{84.15\% $\pm$ 0.33} \\
    \bottomrule
    \end{tabular}
\label{tab:real_human_activity}
\end{table}
  \end{minipage}
  \hfill
  \begin{minipage}[b]{0.45\textwidth}
    \centering
 \begin{table}[H]
    \centering
    \caption{\textbf{Event sequence classification}. Irregular sequential MNIST. Test accuracy (mean $\pm$ std, $N=5$)}
    \vspace{0mm}
    \begin{tabular}{lc}
    \toprule
    Model &  Accuracy \\
    \hline
    ODE-RNN   & 72.41\% $\pm$ 1.69\\
    CT-RNN   & 72.05\% $\pm$ 0.71\\
    Augmented LSTM   & 82.10\% $\pm$ 4.36\\
    CT-GRU  & 87.51\% $\pm$ 1.57\\
    RNN Decay  &  88.93\% $\pm$ 4.06\\
    Bi-directional RNN  & 94.43\% $\pm$ 0.23\\
    GRU-D  & 95.44\% $\pm$ 0.34\\
    PhasedLSTM  & 86.79\% $\pm$ 1.57\\
    GRU-ODE  & 80.95\% $\pm$ 1.52\\ 
    CT-LSTM &  94.84\% $\pm$ 0.17 \\
    \cdashlinelr{1-2}
    ODE-LSTM (ours)  & \textbf{95.73\% $\pm$ 0.24} \\
    \bottomrule
    \end{tabular}
\label{tab:real_mnist}
\end{table}
\end{minipage}
\end{minipage}
\vspace{-8mm}
\end{figure}

This setting realizes a per-time-step classification problem. That is a new error signal is presented to the network at every time-step which makes the vanishing gradient less of an issue here. The results in Table \ref{tab:real_human_activity} shows that the ODE-LSTM outperforms other RNN models on this dataset.
While the significance of an evaluation on a single dataset is limited, it demonstrates that the supreme generalization ability of ODE-LSTM architecture. 


\subsection{Event-based sequential MNIST}
We determined a challenging sequence classification task by designing an event-based version for the sequential-MNIST dataset. For doing this we followed the procedure described below:
\vspace{-\topsep}
\begin{enumerate}
\setlength{\parskip}{0pt}
\setlength{\itemsep}{0pt}
    \item \textbf{Sequentialization + encoding long-term dependencies~~~} transform the 28-by-28 image into a time-series of length 784 
    \item \textbf{Compression + non-uniform sampling~~~} encode binary time-series in a event-based format, to get rid of consecutive occurrences of the same binary value, e.g., $1,1,1,1$ is transformed to $(1,t=4)$. (Read more about this experiment in supplements)
\end{enumerate}
\vspace{-\topsep}

Using this sequentialization mechansim, we compress the sequences from 784 to padded sequences of 256 irregularly-sampled datapoints.
To perform well on this task, RNNs must learn to store some information up to 256 time-steps, while taking the time-lags between them into account. Since an error signal is issued at the end of the sequence, \textit{only an RNN model immune to vanishing gradients can achieve high-degrees of accuracy}.


Table \ref{tab:real_mnist} demonstrates that ODE-based RNN architectures, such as the ODE-RNN, CT-RNN, and the GRU-ODE \cite{de2019gru} struggle to learn a high-fidelity model of this dataset. On the other hand, RNNs built based on a memory mechanism, such as the Bi-directional RNN and GRU-D \cite{che2018recurrent} perform reasonably well, while the performance of ODE-LSTM surpasses that of other models.

\begin{wrapfigure}[8]{r}{0.4\textwidth}
\vspace{-5mm}
\centering
\includegraphics[width=0.4\textwidth]{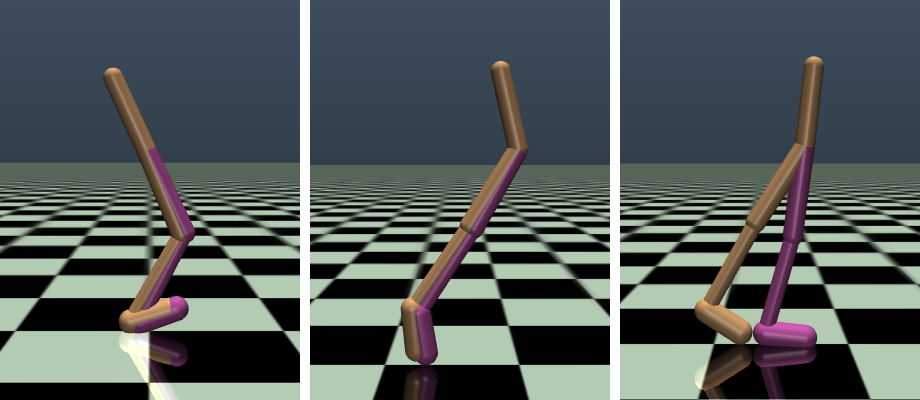}
\vspace{-5mm}
\caption{Walker-2d kinematic dataset}
\label{fig:walker}
\end{wrapfigure}

\subsection{Walker2d kinematic simulation}
In this experiment, we evaluated how well ODE-LSTM can model a physical dynamical system. To this end, we collected simulation data of the \texttt{Walker2d-v2} OpenAI gym \cite{gym} environment using a  pre-trained policy (see Figure \ref{fig:walker}). The objective of the model was to learn the kinematic simulation of the MuJoCo physics engine \cite{todorov2012mujoco} in an auto-regressive fashion and a supervised learning modality.
We increased the complexity of this task by using the pre-trained policy at different training stages (between 500 to 1200 Proximal Policy Optimization (PPO) iterations \cite{schulman2017proximal}) and overwrote 1\% of all actions by random actions.
Moreover, we simulated frame-skips by removing 10\% of the time-steps. Consequently, the dataset is irregularly-sampled. The results, shown in Table \ref{tab:real_walker}, indicate that ODE-LSTM can capture the kinematic dynamics of the physics engine better than other algorithms with a high margin.

\begin{table}[H]
\vspace{-4mm}
    \centering
     \caption{\textbf{Per time-step regression}. Walker2d kinematic dataset. (mean $\pm$ std, $N=5$)}
     \begin{tabular}{lc}
     \toprule
     Model & Square-error \\
\hline
ODE-RNN & 1.904 $\pm$ 0.061 \\
CT-RNN & 1.198 $\pm$ 0.004 \\
Augmented LSTM & 1.065 $\pm$ 0.006 \\
CT-GRU & 1.172 $\pm$ 0.011 \\
RNN-Decay & 1.406 $\pm$ 0.005 \\
Bi-directional RNN & 1.071 $\pm$ 0.009 \\
GRU-D & 1.090 $\pm$ 0.034 \\
PhasedLSTM & 1.063 $\pm$ 0.010 \\
GRU-ODE & 1.051 $\pm$ 0.018 \\
CT-LSTM & 1.014 $\pm$ 0.014 \\
     \cdashlinelr{1-2} 
ODE-LSTM (ours) & \textbf{0.883 $\pm$ 0.014} \\
     \bottomrule
     \end{tabular}
 \label{tab:real_walker}
\end{table}

\section{Discussions, Scope and Limitations}
\textbf{What if we feed in samples' time-lag as an additional input feature to network?} The \textit{Augmented LSTM} architecture we benchmarked against realizes this concept, which is a simplistic approach to making LSTMs compatible with irregularly sampled data. The RNN could then learn to make sense of the time input, for instance, by making its change proportional to the elapsed-time. 

Nonetheless, the time characteristic of an augmented RNN depends purely on its learning process. Consequently, we can only hope that the augmented RNN generalize to unseen time-lags. Our experiments showed that an augmented LSTM performs reasonably well while being outperformed by models that explicitly declare their state by a continuous-time modality, such as ODE-LSTMs.

\textbf{Difference between bidirectional RNNs and ODE-LSTM?~~~} 
A bi-directional architecture consists of two different types of RNNs reciprocally linked together in an auto-regressive fashion \cite{schuster1997bidirectional}. In our context, the first RNN could be designed to handle irregularly-sample time series while the second one is capable of learning long-term dependencies \cite{cao2018brits}. For example, an LSTM bidirectionally coupled with an ODE-RNN could, in principle, overcome both challenges. However, the use of heterogeneous RNN architectures might limit the learning process. In particular, due to different learning speeds, the LSTM could already be overfitting long before the ODE-RNN has learned useful dynamics. 

Contrarily, our ODE-LSTM interlinks LSTMs and ODE-RNNs not in an autoregressive fashion, but at an architectural level, avoiding the problem of learning at different speeds. Our experiments showed that ODE-LSTMs consistently outperform a bi-directional LSTM-ODE-RNN architecture.

\section{Related Works}
\textbf{Time-continuous RNNs~~~} The notion of \textit{CT-RNNs} \cite{funahashi1993approximation} was introduced around three decades ago. It is identical to the ODE-RNN architecture \cite{rubanova2019latent} with an additional dampening factor $\tau$. In our experiments, however, we observed a competitive performance to our ODE-LSTMs achieved by the \textit{GRU-D} architecture \cite{che2018recurrent}. GRU-D encodes the dependence on the time-lags by a trainable decaying mechanism, similar to \textit{RNN-decay} \cite{rubanova2019latent}. While this mechanism enables modeling irregularly sampled time-series, it also introduces a vanishing gradient factor to the backpropagation path. 

Similarly, \textit{CT-GRU} \cite{mozer2017discrete} adds multiple decay factors in the form of extra dimensions to the RNN state. An attention mechanism inside the CT-GRU then selects which entry along the decay dimension to use for computing the next state update. The CT-GRU aims to avoid vanishing gradients by including a decay rate of 0, i.e., no decay at all. This mechanism nevertheless, fails as illustrated in Table \ref{tab:synthetic}. 

\textit{Phased-LSTM} \cite{neil2016phased} adds a learnable oscillator to LSTM. The oscillator modulates LSTM to create dependencies on the elapsed-time, but also introduces a vanishing factor in its gradients. 

\textit{GRU-ODE} \cite{de2019gru} modifies the GRU \cite{chung2014empirical} topology by incorporating a continuous dynamical system. First, GRU is expressed as a discrete difference equation and then transformed into a continuous ODE. This process makes the error-propagation time-dependent, i.e., the near-constant error propagation property of GRU is abolished. 

\textit{CT-LSTM} \cite{mei2017neural} combines the LSTM architecture with continuous-time neural Hawkes processes. At each time-step, the RNN computes two alternative next state options of its hidden state. The actual hidden state is then computed by interpolating between these two hidden states depending on the elapsed time.

\textbf{Learning Irregularly-Sampled Data~~~} Statistical \cite{pearson2003imbalanced,li2016scalable,belletti2016scalable,roy2020robust} and functional analysis \cite{foster1996wavelets,amigo2012transcripts,kowal2019functional} tools have long been studying non-uniformly-spaced data. An alternative and a natural fit for this problem is the use of time-continuous recurrent networks \cite{rubanova2019latent}. We showed that although ODE-RNNs are performant models in these domains, their performance tremendously drops when the incoming samples have long-range dependencies. We solved this shortcoming by introducing ODE-LSTMs. 

\textbf{Learning Long-term Dependencies~~~} The notorious question of vanishing/exploding gradient \cite{hochreiter1991untersuchungen,bengio1994learning} was identified as the core reason for RNNs' lack of generalizability when trained by gradient descent \cite{allen2019can,sherstinsky2020fundamentals}. Recent studies used state-regularization \cite{wang2019state} and long memory stochastic processes \cite{greaves2019statistical} to analyze long-range dependencies. Apart from the original LSTM model \cite{hochreiter1997long} and its variants \cite{greff2016lstm} that solve the problem in the context of RNNs, very few alternative researches exist \cite{chen2019recurrent}. 

As the class of CT RNNs become steadily popularized \cite{hasani2020liquid,lechner2020neural}, it is important to characterize them better \cite{lechner2019designing,dupont2019augmented,durkan2019neural} and understand their applicability and limitations \cite{jia2019neural,lechner2020gershgorin,hanshu2020robustness,holl2020learning,quaglino2020snode,kidger2020neural,hasani2020natural}. In this paper, we proposed a method to enable ODE-based RNNs to learn long-term dependencies.

\section{Conclusion}
We proposed a solution to learn long-term dependencies in irregularly-sampled input data streams. To perform this, we designed a novel long short term memory network, that possesses a continuous-time output state, and consequently modifies its internal dynamical flow to a continuous-time model. ODE-LSTMs resolve the vanishing and exploding of the gradient problem of the class of ODE-RNNs while demonstrating an attractive performance in learning long-term dependencies on data arriving at non-uniform intervals. 
\section*{Broader Impact}

\textbf{Who will benefit from this research?   }
Time series data with missing values and non-uniform intervals are the routine settings in many safety-critical application domains, such as medical, business, social, and the automation of industries.

The results of this paper enable users to construct learning systems that not only help handle irregularly sampled data efficiently but also to learn long-term dependencies that might be vital to their application. 

For instance, consider the decision-critical domain of surgical processes or the treatment of patients in intensive care units (ICU) in which the medical team has to have access to the process actively, and the steps are taken throughout a surgical procedure, to make/take a current decision/action. An intelligent agent in use as an assistant during surgery must be able to do the same and carefully assign credits to the actions taken in the past (long-term dependencies) to output an accurate decision. This example simultaneously consists of irregularly-sampled inputs and long-term dependencies. Our proposed method enables these modalities. 

\textbf{Preventing failure of the system   } Like any other intelligent system, our proposed algorithm has to go through robustness analysis (perturbations, noise, and adversarial attack), before being deployed in high-stakes decision-making applications. This process would dramatically reduce the chance of failure of intelligent systems such as ours. 
 
\textbf{Whether the method leverages biases in the data   } The mechanisms of "learning to forget" and "learning long-term dependencies" are encoded in our proposed method. Both processes can be used as the controller of biases in data, and help us design \emph{fair} machine learning systems.

\begin{ack}
M.L. is supported in parts by the Austrian Science Fund (FWF) under grant Z211-N23 (Wittgenstein Award). R.H. is partially supported by the Horizon-2020 ECSEL Project grant No. 783163 (iDev40), and Boeing.
\end{ack}


\bibliographystyle{plain}
\bibliography{references}

\begin{thebibliography}{10}

\bibitem{tensorflow2015-whitepaper}
Mart\'{\i}n Abadi, Ashish Agarwal, Paul Barham, Eugene Brevdo, Zhifeng Chen,
  Craig Citro, Greg~S. Corrado, Andy Davis, Jeffrey Dean, Matthieu Devin,
  Sanjay Ghemawat, Ian Goodfellow, Andrew Harp, Geoffrey Irving, Michael Isard,
  Yangqing Jia, Rafal Jozefowicz, Lukasz Kaiser, Manjunath Kudlur, Josh
  Levenberg, Dan Man\'{e}, Rajat Monga, Sherry Moore, Derek Murray, Chris Olah,
  Mike Schuster, Jonathon Shlens, Benoit Steiner, Ilya Sutskever, Kunal Talwar,
  Paul Tucker, Vincent Vanhoucke, Vijay Vasudevan, Fernanda Vi\'{e}gas, Oriol
  Vinyals, Pete Warden, Martin Wattenberg, Martin Wicke, Yuan Yu, and Xiaoqiang
  Zheng.
\newblock {TensorFlow}: Large-scale machine learning on heterogeneous systems,
  2015.
\newblock Software available from tensorflow.org.

\bibitem{allen2019can}
Zeyuan Allen-Zhu and Yuanzhi Li.
\newblock Can sgd learn recurrent neural networks with provable generalization?
\newblock In {\em Advances in Neural Information Processing Systems}, pages
  10331--10341, 2019.

\bibitem{amigo2012transcripts}
Jos{\'e}~M Amig{\'o}, Roberto Monetti, Thomas Aschenbrenner, and Wolfram Bunk.
\newblock Transcripts: An algebraic approach to coupled time series.
\newblock {\em Chaos: An Interdisciplinary Journal of Nonlinear Science},
  22(1):013105, 2012.

\bibitem{belletti2016scalable}
Francois~W Belletti, Evan~R Sparks, Michael~J Franklin, Alexandre~M Bayen, and
  Joseph~E Gonzalez.
\newblock Scalable linear causal inference for irregularly sampled time series
  with long range dependencies.
\newblock {\em arXiv preprint arXiv:1603.03336}, 2016.

\bibitem{bengio1994learning}
Yoshua Bengio, Patrice Simard, and Paolo Frasconi.
\newblock Learning long-term dependencies with gradient descent is difficult.
\newblock {\em IEEE transactions on neural networks}, 5(2):157--166, 1994.

\bibitem{gym}
Greg Brockman, Vicki Cheung, Ludwig Pettersson, Jonas Schneider, John Schulman,
  Jie Tang, and Wojciech Zaremba.
\newblock Openai gym, 2016.

\bibitem{cao2018brits}
Wei Cao, Dong Wang, Jian Li, Hao Zhou, Lei Li, and Yitan Li.
\newblock Brits: Bidirectional recurrent imputation for time series.
\newblock In {\em Advances in Neural Information Processing Systems}, pages
  6775--6785, 2018.

\bibitem{che2018recurrent}
Zhengping Che, Sanjay Purushotham, Kyunghyun Cho, David Sontag, and Yan Liu.
\newblock Recurrent neural networks for multivariate time series with missing
  values.
\newblock {\em Scientific reports}, 8(1):1--12, 2018.

\bibitem{chen2019recurrent}
Dexiong Chen, Laurent Jacob, and Julien Mairal.
\newblock Recurrent kernel networks.
\newblock In {\em Advances in Neural Information Processing Systems}, pages
  13431--13442, 2019.

\bibitem{chen2018neural}
Tian~Qi Chen, Yulia Rubanova, Jesse Bettencourt, and David~K Duvenaud.
\newblock Neural ordinary differential equations.
\newblock In {\em Advances in neural information processing systems}, pages
  6571--6583, 2018.

\bibitem{chung2014empirical}
Junyoung Chung, Caglar Gulcehre, KyungHyun Cho, and Yoshua Bengio.
\newblock Empirical evaluation of gated recurrent neural networks on sequence
  modeling.
\newblock {\em arXiv preprint arXiv:1412.3555}, 2014.

\bibitem{de2019gru}
Edward De~Brouwer, Jaak Simm, Adam Arany, and Yves Moreau.
\newblock Gru-ode-bayes: Continuous modeling of sporadically-observed time
  series.
\newblock In {\em Advances in Neural Information Processing Systems}, pages
  7377--7388, 2019.

\bibitem{dormand1980family}
John~R Dormand and Peter~J Prince.
\newblock A family of embedded runge-kutta formulae.
\newblock {\em Journal of computational and applied mathematics}, 6(1):19--26,
  1980.

\bibitem{dua2019}
Dheeru Dua and Casey Graff.
\newblock {UCI} machine learning repository, 2017.

\bibitem{dupont2019augmented}
Emilien Dupont, Arnaud Doucet, and Yee~Whye Teh.
\newblock Augmented neural odes.
\newblock In {\em Advances in Neural Information Processing Systems}, pages
  3134--3144, 2019.

\bibitem{durkan2019neural}
Conor Durkan, Artur Bekasov, Iain Murray, and George Papamakarios.
\newblock Neural spline flows.
\newblock In {\em Advances in Neural Information Processing Systems}, pages
  7509--7520, 2019.

\bibitem{foster1996wavelets}
Grant Foster.
\newblock Wavelets for period analysis of unevenly sampled time series.
\newblock {\em The Astronomical Journal}, 112:1709, 1996.

\bibitem{funahashi1993approximation}
Ken-ichi Funahashi and Yuichi Nakamura.
\newblock Approximation of dynamical systems by continuous time recurrent
  neural networks.
\newblock {\em Neural networks}, 6(6):801--806, 1993.

\bibitem{gers1999learning}
Felix~A Gers, J{\"u}rgen Schmidhuber, and Fred Cummins.
\newblock Learning to forget: Continual prediction with lstm.
\newblock 1999.

\bibitem{gholami2019anode}
Amir Gholami, Kurt Keutzer, and George Biros.
\newblock Anode: Unconditionally accurate memory-efficient gradients for neural
  odes.
\newblock {\em arXiv preprint arXiv:1902.10298}, 2019.

\bibitem{greaves2019statistical}
Alexander Greaves-Tunnell and Zaid Harchaoui.
\newblock A statistical investigation of long memory in language and music.
\newblock In {\em International Conference on Machine Learning}, pages
  2394--2403, 2019.

\bibitem{greff2016lstm}
Klaus Greff, Rupesh~K Srivastava, Jan Koutn{\'\i}k, Bas~R Steunebrink, and
  J{\"u}rgen Schmidhuber.
\newblock Lstm: A search space odyssey.
\newblock {\em IEEE transactions on neural networks and learning systems},
  28(10):2222--2232, 2016.

\bibitem{hanshu2020robustness}
YAN Hanshu, DU~Jiawei, TAN Vincent, and FENG Jiashi.
\newblock On robustness of neural ordinary differential equations.
\newblock In {\em International Conference on Learning Representations}, 2020.

\bibitem{hasani2020liquid}
Ramin Hasani, Mathias Lechner, Alexander Amini, Daniela Rus, and Radu Grosu.
\newblock Liquid time-constant networks.
\newblock {\em arXiv preprint arXiv:2006.04439}, 2020.

\bibitem{hasani2020natural}
Ramin Hasani, Mathias Lechner, Alexander Amini, Daniela Rus, and Radu Grosu.
\newblock The natural lottery ticket winner: Reinforcement learning with
  ordinary neural circuits.
\newblock In {\em Proceedings of the 2020 International Conference on Machine
  Learning}. JMLR. org, 2020.

\bibitem{hochreiter1991untersuchungen}
Sepp Hochreiter.
\newblock Untersuchungen zu dynamischen neuronalen netzen [in german] diploma
  thesis.
\newblock {\em TU M{\"u}nich}, 1991.

\bibitem{hochreiter1997long}
Sepp Hochreiter and J{\"u}rgen Schmidhuber.
\newblock Long short-term memory.
\newblock {\em Neural computation}, 9(8):1735--1780, 1997.

\bibitem{holl2020learning}
Philipp Holl, Vladlen Koltun, and Nils Thuerey.
\newblock Learning to control pdes with differentiable physics.
\newblock {\em arXiv preprint arXiv:2001.07457}, 2020.

\bibitem{jia2019neural}
Junteng Jia and Austin~R Benson.
\newblock Neural jump stochastic differential equations.
\newblock In {\em Advances in Neural Information Processing Systems}, pages
  9843--9854, 2019.

\bibitem{kidger2020neural}
Patrick Kidger, James Morrill, James Foster, and Terry Lyons.
\newblock Neural controlled differential equations for irregular time series.
\newblock {\em arXiv preprint arXiv:2005.08926}, 2020.

\bibitem{kowal2019functional}
Daniel~R Kowal, David~S Matteson, and David Ruppert.
\newblock Functional autoregression for sparsely sampled data.
\newblock {\em Journal of Business \& Economic Statistics}, 37(1):97--109,
  2019.

\bibitem{lechner2020neural}
Mathias Lechner, Ramin Hasani, Alexander Amini, Thomas~A Henzinger, Daniela
  Rus, and Radu Grosu.
\newblock Neural circuit policies enabling auditable autonomy.
\newblock {\em Nature Machine Intelligence}, 2(10):642--652, 2020.

\bibitem{lechner2020gershgorin}
Mathias Lechner, Ramin Hasani, Daniela Rus, and Radu Grosu.
\newblock Gershgorin loss stabilizes the recurrent neural network compartment
  of an end-to-end robot learning scheme.
\newblock In {\em 2020 International Conference on Robotics and Automation
  (ICRA)}. IEEE, 2020.

\bibitem{lechner2019designing}
Mathias Lechner, Ramin Hasani, Manuel Zimmer, Thomas~A Henzinger, and Radu
  Grosu.
\newblock Designing worm-inspired neural networks for interpretable robotic
  control.
\newblock In {\em 2019 International Conference on Robotics and Automation
  (ICRA)}, pages 87--94. IEEE, 2019.

\bibitem{lecun1998gradient}
Yann LeCun, L{\'e}on Bottou, Yoshua Bengio, and Patrick Haffner.
\newblock Gradient-based learning applied to document recognition.
\newblock {\em Proceedings of the IEEE}, 86(11):2278--2324, 1998.

\bibitem{li2016scalable}
Steven Cheng-Xian Li and Benjamin~M Marlin.
\newblock A scalable end-to-end gaussian process adapter for irregularly
  sampled time series classification.
\newblock In {\em Advances in neural information processing systems}, pages
  1804--1812, 2016.

\bibitem{liang2018rllib}
Eric Liang, Richard Liaw, Robert Nishihara, Philipp Moritz, Roy Fox, Ken
  Goldberg, Joseph Gonzalez, Michael Jordan, and Ion Stoica.
\newblock Rllib: Abstractions for distributed reinforcement learning.
\newblock In {\em International Conference on Machine Learning}, pages
  3053--3062, 2018.

\bibitem{marvin1969perceptrons}
Minsky Marvin and A~Papert Seymour.
\newblock {\em Perceptrons}.
\newblock MIT Press, 1969.

\bibitem{mei2017neural}
Hongyuan Mei and Jason~M Eisner.
\newblock The neural hawkes process: A neurally self-modulating multivariate
  point process.
\newblock In {\em Advances in Neural Information Processing Systems}, pages
  6754--6764, 2017.

\bibitem{mozer2017discrete}
Michael~C Mozer, Denis Kazakov, and Robert~V Lindsey.
\newblock Discrete event, continuous time rnns.
\newblock {\em arXiv preprint arXiv:1710.04110}, 2017.

\bibitem{neil2016phased}
Daniel Neil, Michael Pfeiffer, and Shih-Chii Liu.
\newblock Phased lstm: Accelerating recurrent network training for long or
  event-based sequences.
\newblock In {\em Advances in neural information processing systems}, pages
  3882--3890, 2016.

\bibitem{pascanu2013difficulty}
Razvan Pascanu, Tomas Mikolov, and Yoshua Bengio.
\newblock On the difficulty of training recurrent neural networks.
\newblock In {\em International conference on machine learning}, pages
  1310--1318, 2013.

\bibitem{NEURIPS2019_9015}
Adam Paszke, Sam Gross, Francisco Massa, Adam Lerer, James Bradbury, Gregory
  Chanan, Trevor Killeen, Zeming Lin, Natalia Gimelshein, Luca Antiga, Alban
  Desmaison, Andreas Kopf, Edward Yang, Zachary DeVito, Martin Raison, Alykhan
  Tejani, Sasank Chilamkurthy, Benoit Steiner, Lu~Fang, Junjie Bai, and Soumith
  Chintala.
\newblock Pytorch: An imperative style, high-performance deep learning library.
\newblock In H.~Wallach, H.~Larochelle, A.~Beygelzimer, F.~dAlch\'{e} Buc,
  E.~Fox, and R.~Garnett, editors, {\em Advances in Neural Information
  Processing Systems 32}, pages 8024--8035. Curran Associates, Inc., 2019.

\bibitem{pearson2003imbalanced}
Ronald Pearson, Gregory Goney, and James Shwaber.
\newblock Imbalanced clustering for microarray time-series.
\newblock In {\em Proceedings of the ICML}, volume~3, 2003.

\bibitem{pontryagin2018mathematical}
Lev~Semenovich Pontryagin.
\newblock {\em Mathematical theory of optimal processes}.
\newblock Routledge, 2018.

\bibitem{quaglino2020snode}
Alessio Quaglino, Marco Gallieri, Jonathan Masci, and Jan Koutník.
\newblock Snode: Spectral discretization of neural odes for system
  identification.
\newblock In {\em International Conference on Learning Representations}, 2020.

\bibitem{roy2020robust}
DP~Roy and L~Yan.
\newblock Robust landsat-based crop time series modelling.
\newblock {\em Remote Sensing of Environment}, 238:110810, 2020.

\bibitem{rubanova2019latent}
Yulia Rubanova, Tian~Qi Chen, and David~K Duvenaud.
\newblock Latent ordinary differential equations for irregularly-sampled time
  series.
\newblock In {\em Advances in Neural Information Processing Systems}, pages
  5321--5331, 2019.

\bibitem{rumelhart1986learning}
David~E Rumelhart, Geoffrey~E Hinton, and Ronald~J Williams.
\newblock Learning representations by back-propagating errors.
\newblock {\em nature}, 323(6088):533--536, 1986.

\bibitem{runge1895numerische}
Carl Runge.
\newblock {\"U}ber die numerische aufl{\"o}sung von differentialgleichungen.
\newblock {\em Mathematische Annalen}, 46(2):167--178, 1895.

\bibitem{schulman2017proximal}
John Schulman, Filip Wolski, Prafulla Dhariwal, Alec Radford, and Oleg Klimov.
\newblock Proximal policy optimization algorithms.
\newblock {\em arXiv preprint arXiv:1707.06347}, 2017.

\bibitem{schuster1997bidirectional}
Mike Schuster and Kuldip~K Paliwal.
\newblock Bidirectional recurrent neural networks.
\newblock {\em IEEE transactions on Signal Processing}, 45(11):2673--2681,
  1997.

\bibitem{sherstinsky2020fundamentals}
Alex Sherstinsky.
\newblock Fundamentals of recurrent neural network (rnn) and long short-term
  memory (lstm) network.
\newblock {\em Physica D: Nonlinear Phenomena}, 404:132306, 2020.

\bibitem{tieleman2012lecture}
Tijmen Tieleman and Geoffrey Hinton.
\newblock Lecture 6.5-rmsprop: Divide the gradient by a running average of its
  recent magnitude.
\newblock {\em COURSERA: Neural networks for machine learning}, 4(2):26--31,
  2012.

\bibitem{todorov2012mujoco}
Emanuel Todorov, Tom Erez, and Yuval Tassa.
\newblock Mujoco: A physics engine for model-based control.
\newblock In {\em 2012 IEEE/RSJ International Conference on Intelligent Robots
  and Systems}, pages 5026--5033. IEEE, 2012.

\bibitem{wang2019state}
Cheng Wang and Mathias Niepert.
\newblock State-regularized recurrent neural networks.
\newblock In {\em International Conference on Machine Learning}, pages
  6596--6606, 2019.

\bibitem{werbos1988generalization}
Paul~J Werbos.
\newblock Generalization of backpropagation with application to a recurrent gas
  market model.
\newblock {\em Neural networks}, 1(4):339--356, 1988.

\bibitem{werbos1990backpropagation}
Paul~J Werbos.
\newblock Backpropagation through time: what it does and how to do it.
\newblock {\em Proceedings of the IEEE}, 78(10):1550--1560, 1990.

\bibitem{zhuang2020adaptive}
Juntang Zhuang, Nicha Dvornek, Xiaoxiao Li, Sekhar Tatikonda, Xenophon
  Papademetris, and James Duncan.
\newblock Adaptive checkpoint adjoint method for gradient estimation in neural
  ode.
\newblock In {\em Proceedings of the 37th International Conference on Machine
  Learning}. PMLR 119, 2020.

\end{thebibliography}

\clearpage

\beginsupplement

\text{\huge \textbf{Supplementary Materials}}

\section{Proofs}
\textbf{Derivation of the Euler's method Jacobian}
Let $\dot{h} = f_\theta(x,h,T) - h \tau$ be an ODE-RNN. 
Then the explicit Euler's method with step-size $T$ is defined as the discretization 
\begin{equation}
h_{t+T} = h_t + T (f_\theta(x,h,T)-h\tau)\Big|_{h=h_t}.
\end{equation}
Therefore, state-previous state Jacobian is given by

    \begin{equation}
         \frac{\partial h_{t+T}}{\partial h_t} = I + T\frac{\partial f}{\partial h}\Big|_{h=h_t} - \tau T I.
    \end{equation}

\textbf{Derivation of the Runge-Kutta Jacobian}
Let $\dot{h} = f_\theta(x,h,T) - h \tau$ be an ODE-RNN. 
Then the Runge-Kutta method with step-size $T$ is defined as the discretization 
\begin{equation}
h_{t+T} = h_t + T \sum_{j=1}^{M} b_i (f_\theta(x,h,T)-h\tau)\Big|_{h=K_i},
\end{equation}
where the coefficients $b_i$ and the values $K_i$ are taken according to the Butcher tableau with $\sum_{j=1}^{M} b_i=1$ and $K_1=h_t$.

Then state-previous state Jacobian of the Runge-Kutta method  is given by the following equation :
    \begin{equation}
          \frac{\partial h_{t+T}}{\partial h_t} = I + T\sum_{j=1}^{M} b_i \frac{\partial f}{\partial h}\Big|_{h=K_i} - \tau T I.,
    \end{equation}

Note that the explicit Euler method is an instance of the Runge-Kutta method with $M=1$ and $b_1=1$.

\textbf{Proof of ODE-RNN suffering from vanishing or exploding gradients}
Let $\dot{h} = f_\theta(x,h,T) - h \tau$ be an ODE-RNN with latent dimension $N$. Without loss of generality let $h_0$ be the initial state at $t=0$ and $h_T$ denote the ODE state which should be computed by a numerical ODE-solver.
Then ODE-solvers, including fixed-step methods \cite{runge1895numerische} and variable-step methods such as the Dormand-Prince method \cite{dormand1980family}, discretize the interval $[0,T]$ by a series $t_0,t_1,\dots t_n$, where $t_0=0$ and $t_n=T$ and each $h_{t_i}$ is computed by a single-step explicit Euler or Runge-Kutta method from $h_{t_{i-1}}$. 

Our proof closely aligns with the analysis in Hochreiter and Schmidhuber \cite{hochreiter1997long}. We refer the reader to \cite{hochreiter1991untersuchungen,bengio1994learning,pascanu2013difficulty} for a rigorous discussion on the vanishing and exploding gradients. 

We first prove the theorem for a scalar RNN, i.e., $n=1$, and then extend the discussion to the general case.
The error-flow per RNN step between $t=0$ and $t=T$ is given by
\begin{equation}
    \frac{\partial h_{T}}{\partial h_{0}} = \prod_{m=1}^{n}\Big( 1 + (t_m-t_{m-1})\sum_{j=1}^{M} b_i \frac{\partial f}{\partial h}\Big|_{h=K_{m_i}} - \tau (t_m-t_{m-1}) \Big),
\end{equation}
which realizes a power series depending on the value 
\begin{equation}\label{eq:abs_value}
    | 1 + (t_m-t_{m-1})\sum_{j=1}^{M} b_i \frac{\partial f}{\partial h}\Big|_{h=K_{m_i}} - \tau (t_m-t_{m-1})|.
\end{equation}
Obviously, the condition that this term is equal to 1 is not enforced during training and violated for any non-trivial $f_\theta$, such as $f_{\theta}(h,x) = \sigma(W_h h + W_x x + \hat{b})$ with $\sigma$ being a sigmoidal or rectified-linear activation function. The exact magnitude depends on the weights $W_h$, as
\begin{equation}
    \frac{\partial f_{\theta}(h,x)}{\partial h} = W_h \sigma'(W_h h + W_x x + \hat{b}).
\end{equation}
A non-zero time-constant $\tau$ pushes the gradient toward a vanishing region.

Note that the Equation (\ref{eq:abs_value}) only becomes equal to 1, if $\sum_{j=1}^{M} b_i \frac{\partial f}{\partial h}\Big|_{h=K_{m_i}}=\tau$. This would imply that $\frac{\partial h_{t_m}}{h_{t_m-1}} =0$, i.e., when the change in ODE-state between two time-points is zero. A variable that does not change over time is a memory element. Thus the only solution of enforcing a constant-error propagation is to include an explicit memory element in the architecture \cite{hochreiter1997long} which does not change its value between two arbitrary time-points $t_m$ and $t_{m-1}$.

For the general case $n\geq 1$, the error-flow per RNN step between $t=0$ and $t=T$ is given by
\begin{equation}
    \frac{\partial h_{T}}{\partial h_{0}} = \prod_{m=1}^{n}\Big( I + (t_m-t_{m-1})\sum_{j=1}^{M} b_i \frac{\partial f}{\partial h}\Big|_{h=K_{m_i}} - \tau (t_m-t_{m-1}) I \Big).
\end{equation}
As $h$ is a vector, we need to consider all possible error-propagation paths.
The error-flow from unit $u$ to unit $v$ is then given by summing all $N^{n-1}$ possible paths between $u$ to $v$,
\begin{equation}\label{eq:big}
    \frac{\partial h^v_{T}}{\partial h^u_{0}} =  \sum_{l_1}^{N} \dots \sum_{l_{n-1}}^{N} \prod_{m=1}^{n}\Big( I + (t_m-t_{m-1})\sum_{j=1}^{M} b_i \frac{\partial f}{\partial h}\Big|_{h=K_{m_i}} - \tau (t_m-t_{m-1}) I \Big)_{l_m,l_{m-1}},
\end{equation}
where $l_0=u$ and $l_n=v$.

The arguments of the scalar case hold for every individual path in Equation (\ref{eq:big}).
The only difference between the the scalar case and the individual paths in the vectored version is the non-diagonal connections in the general case do not include the constant 1 and $\tau$. The error-propagation magnitude between $u$ and $v$ with $u\neq v$ is given by 
\begin{equation}\label{eq:nondiag_value}
    | (t_m-t_{m-1})\Big(\sum_{j=1}^{M} b_i \frac{\partial f}{\partial h}\Big|_{h=K_{m_i}} \Big)_{u,v} |.
\end{equation}

Again, for $f_{\theta}(h,x) = \sigma(W_h h + W_x x + \hat{b})$ we obtain an error-flow that depends on the weights $W_h$ and can be either vanishing or exploding, depending on its magnitude.

\textbf{Proof that even gradients of the ODE solution can vanish or explode}
Let $\dot{h} = f_\theta(x,h,T) - h \tau$ be an ODE-RNN with latent dimension $N$, with $f_\theta$ being uniformly Lipschitz continuous. Without loss of generality let $h_0$ be the initial state at $t=0$ and $h_T$ denote the ODE state which should be computed by a numerical ODE-solver.
We approximate the interval $[0,T]$ by a uniform discretization grid, i.e. $t_i-t_{i-1} = t_j-t_{j-1} = T/n$ for all $i,j$ $t_0,t_1,\dots t_n$, where $t_0=0$ and $t_n=T$ and each $h_{t_i}$ is computed by a single-step explicit Euler from $h_{t_{i-1}}$. 

Even when making the discretization grid $t_0,t_1,\dots t_n$ finer and finer, the gradient propagation issue is not resolved. Let $h_i$ denote the intermediate values computed by the Picard-iteration, i.e., the explicit Euler. By the Picard–Lindelöf theorem, we know that $h_T$ converges to the true solution $h(T)$. 

First, we assume there exists a $\xi>0$ such that  $\xi \leq \frac{\partial f}{\partial h}\Big|_{h=h_{m}} - \tau\text { for all }m$. Note that this situation can naturally occur if we have a  $f_{\theta}(h,x) = \sigma(W_h h + W_x x + \hat{b})$.
In the limit $n\rightarrow \infty$ we get
\begin{align*}
\lim_{n\rightarrow \infty} \frac{\partial h_{T}}{\partial h_{0}} & = \lim_{n\rightarrow \infty}  \prod_{m=1}^{n}\Big( 1 + (t_m-t_{m-1}) \frac{\partial f}{\partial h}\Big|_{h=h_{m}} - \tau (t_m-t_{m-1}) \Big)\\
&=  \lim_{n\rightarrow \infty}  \prod_{m=1}^{n}\Big( 1 + \frac{T}{n} \frac{\partial f}{\partial h}\Big|_{h=h_{m}} - \tau \frac{T}{n}) \Big)\\
&\geq  \lim_{n\rightarrow \infty}  \prod_{m=1}^{n}\Big( 1 + \frac{T}{n} \xi \Big), \text{  with some  } 0 < \xi \leq \frac{\partial f}{\partial h}\Big|_{h=h_{m}} - \tau\text { for all }m\\
&=  \lim_{n\rightarrow \infty}  \Big( 1 + \frac{T}{n} \xi \Big)^n \\
& = e^{T\xi}\\
& > 1,
\end{align*}
i.e., we have an exploding gradient.

Conversely, lets assume there exists a $\xi<0$ such that  $\xi \geq \frac{\partial f}{\partial h}\Big|_{h=h_{m}} - \tau\text { for all }m$. Note that this situation can also naturally occur, for instance if $\tau>0$ and regions where $f'$ is small.
In the limit $n\rightarrow \infty$ we get
\begin{align*}
\lim_{n\rightarrow \infty} \frac{\partial h_{T}}{\partial h_{0}} & = \lim_{n\rightarrow \infty}  \prod_{m=1}^{n}\Big( 1 + (t_m-t_{m-1}) \frac{\partial f}{\partial h}\Big|_{h=h_{m}} - \tau (t_m-t_{m-1}) \Big)\\
&=  \lim_{n\rightarrow \infty}  \prod_{m=1}^{n}\Big( 1 + \frac{T}{n} \frac{\partial f}{\partial h}\Big|_{h=h_{m}} - \tau \frac{T}{n}) \Big)\\
&\leq  \lim_{n\rightarrow \infty}  \prod_{m=1}^{n}\Big( 1 + \frac{T}{n} \xi \Big), \text{  with some  } 0 > \xi \geq \frac{\partial f}{\partial h}\Big|_{h=h_{m}} - \tau\text { for all }m\\
&=  \lim_{n\rightarrow \infty}  \Big( 1 + \frac{T}{n} \xi \Big)^n \\
& = e^{T\xi}\\
& < 1,
\end{align*}
i.e., we have a vanishing gradient.

Similar to the argument in the proof above, we can extend the scalar case to the general case.
However, summing over all possible path might not be trivial, as the number of possible path also growths to infinity.
\begin{equation}\label{eq:big2}
\lim_{n\rightarrow \infty} \frac{\partial h^v_{T}}{\partial h^u_{0}} = \lim_{n\rightarrow \infty}  \sum_{l_1}^{N} \dots \sum_{l_{n-1}}^{N} \prod_{m=1}^{n}\Big( I + (t_m-t_{m-1}) \frac{\partial f}{\partial h}\Big|_{h=h_{m}} - \tau (t_m-t_{m-1}) I \Big)_{l_m,l_{m-1}}.
\end{equation}
Instead, we assume $u=v=l_1=\dots l_n-1$, i.e., we only look at the error-propagation through the diagonal element $u$.
\begin{align*}
\lim_{n\rightarrow \infty} \frac{\partial h^v_{T}}{\partial h^u_{0}} &= \lim_{n\rightarrow \infty}  \prod_{m=1}^{n}\Big( I + (t_m-t_{m-1})\frac{\partial f}{\partial h}\Big|_{h=h_{m}} - \tau (t_m-t_{m-1}) I \Big)_{u,u}\\
 & = \lim_{n\rightarrow \infty}  \prod_{m=1}^{n}\Big( 1 + (t_m-t_{m-1}) \frac{\partial f^u}{\partial h^u}\Big|_{h^u=h^u_{m}} - \tau^u (t_m-t_{m-1}) \Big)\\
&=  \lim_{n\rightarrow \infty}  \prod_{m=1}^{n}\Big( 1 + \frac{T}{n} \frac{\partial f^u}{\partial h^u}\Big|_{h^u=h^u_{m}} - \tau^u \frac{T}{n}) \Big),
\end{align*}
which is equivalent to the scalar case.
For an interesting $f$ such as $f_{\theta}(h,x) = \sigma(W_h h + W_x x + \hat{b})$, the term $\frac{f}{h}$ depends on the value $W^{u,u}_h$.
By assuming $W^{w,z}$ for any $(w,z)\neq (u,u)$ is neglectable small, we can infer that the effects of the gradient by any other path in Equation (\ref{eq:big2}) is neglectable small. Thus the global error flow depends on $W^{u,u}_h$, which can make the error-flow either explode or vanish depending on its value.

Note that this argument is similar to arguing that as the multi-dimensional case properly contains the scalar case, the multi-dimensional case can express an exploding or vanishing gradient too.

\textbf{Proof that the ODE-LSTM does not suffer from a vanishing or exploding gradient}

Recall that we assume that $R_z, R_i, R_f, W_f$ and $b_f$ are initialized close to 0 and that we are at the beginning of the training process, i.e., we assume the weights do not differ significantly from their initialized values.

We have
\begin{align*}
\frac{\partial c_{t+1}}{\partial c_t} &= \frac{\partial z_{t+1} \elmentwisemul i_{t+1} + c_{t}\elmentwisemul f_{t+1}}{\partial c_t} \\
&= \frac{\partial z_{t+1}}{\partial c_t}  \text{diag}(i_{t+1}) +  \frac{\partial  i_{t+1}}{\partial c_t}  \text{diag}(z_{t+1}) + \text{diag}(f_{t+1}) +  \frac{\partial  f_{t+1}}{\partial c_t} \text{diag}(c_t).
\end{align*}


For the derivative of the input update activation we can simply apply the chain-rule and get
\begin{align*}
    \frac{\partial z_{t+1}^v}{\partial c_t^u} &= \tanh'(W_z x_{t+1} + R_z h_{t} + b_z)^v R_z^{u,v} \frac{\partial h_t^u}{\partial c_t^u}\\
     &= \tanh'(W_z x_{t+1} + R_z h_{t} + b_z)^v R_z^{u,v} o_t^u,
\end{align*}
where $\tanh'$ denotes the functional derivative of the hyperbolic tangent.
As $0 \leq \tanh' \leq 1$,  $0 \leq o_t^u \leq 1$ and most importantly $R_z$ is initialized close to 0, we can safely assume that
\begin{align*}
 \frac{\partial z_{t+1}^v}{\partial c_t^u}  \approx 0.
\end{align*}

Similar argument holds for the input and forget gate derivatives, where we assumed that $R_i$ and $R_f$ are initialized close to 0. Therefore
\begin{align*}
    \frac{\partial i_{t+1}^v}{\partial c_t^u} &= \sigma'(W_i x_{t+1} + R_i h_{t} + b_i)^v R_i^{u,v} o_t^u\\
    & \approx 0
\end{align*}
and
\begin{align*}
    \frac{\partial f_{t+1}^v}{\partial c_t^u} &= \sigma'(W_f x_{t+1} + R_f h_{t} + b_f + \mathbf{1})^v R_f^{u,v} o_t^{u}\\
    & \approx 0,
\end{align*}
where $\sigma'$ denotes the functional derivatives of the sigmoid function.

Consequently, with a proper weight initialization, the Jacobian simplifies to 
\begin{align*}
\frac{\partial c_{t+1}}{\partial c_t} &\approx \text{diag}(f_{t+1}).
\end{align*}

We assumed that $W_f$ and $b_f$ are initialized close to 0. Hence,
\begin{align*}
    f_{t+t}^v & = \sigma(W_f x_{t+1} + R_f h_{t} + b_f + \mathbf{1})^v \\
    & \approx \sigma(1) \\
    & \approx 0.7310586.
\end{align*}

Hence, we have 
\begin{align*}
\Big| \sum_{j=1}^{N} \frac{\partial c_{t+1}^i}{\partial c_t^j}\Big| & \approx 0.7310586,
\end{align*}
,  which is less than 1 (no exploding) but much greater than 0 (no vanishing) and 
ensures a near-constant error propagation at the beginning of the training process.

As already mentioned in the paper, the exact value of the error flow can be controlled by changing the forget gate bias from its default value of 1.
If the underlying data distribution contains dependencies with a very long time-lag, we can bring the error flow factor closer to 1 by increasing forget gate bias. Thus enabling the ODE-LSTM to learn even very long-term dependencies in the data.

\section{Experimental evaluation}
For models containing differential equations, we used the ODE-solvers as listed in Table \ref{tab:ode_solvers}. Hyperparameter settings used for our evaluation is shown in Table \ref{tab:hyperparams}. 

\textbf{Batching}
Sequences of our event-based bit-stream classification task and event-based seqMNIST can have different lengths.
To allow an arbitrary batching of several sequences, we pad all sequences to equal length and apply a binary mask during training and evaluation.

\subsection{Dataset description}
The individual datasets are created as follows:

 \begin{figure}
     \centering
     \begin{subfigure}[t]{0.49\textwidth}
         \centering
         \begin{tikzpicture}[samplea/.style={inner sep=0.5mm,circle,fill=red},sampleb/.style={inner sep=0.5mm,circle,fill=blue},txta/.style={color=red},txtb/.style={color=blue}]
         \draw[-Latex] (0,0) to (5,0);
         \node at (5,-0.4) {Time};
         \draw (0.5,-0.1) -- +(0,0.2);
         \draw (1.0,-0.1) -- +(0,0.2);
         \draw (1.5,-0.1) -- +(0,0.2);
         \draw (2.0,-0.1) -- +(0,0.2);
         \draw (2.5,-0.1) -- +(0,0.2);
         \draw (3.0,-0.1) -- +(0,0.2);
         \draw (3.5,-0.1) -- +(0,0.2);
         \draw (4.0,-0.1) -- +(0,0.2);
         \node (s1) at (0.5,1.2) [samplea] {};
         \node (s2) at (1.0,1.2) [samplea] {};
         \node (s3) at (1.5,1.2) [samplea] {};
         \node (s4) at (2.0,1.2) [samplea] {};
         \node (s5) at (2.5,0.4) [sampleb] {};
         \node (s6) at (3.0,0.4) [sampleb] {};
         \node (s7) at (3.5,0.4) [sampleb] {};
         \node (s8) at (4.0,1.2) [samplea] {};
       \draw[opacity=0.2] (s1) -- +(0,-1.2);
         \draw[opacity=0.2] (s2) -- +(0,-1.2);
        \draw[opacity=0.2] (s3) -- +(0,-1.2);
         \draw[opacity=0.2] (s4) -- +(0,-1.2);
         \draw[opacity=0.2] (s5) -- +(0,-0.4);
         \draw[opacity=0.2] (s6) -- +(0,-0.4);
         \draw[opacity=0.2] (s7) -- +(0,-0.4);
         \draw[opacity=0.2] (s8) -- +(0,-1.2);
         \node (s1) at (0.5,1.5) [txta] {a};
         \node (s1) at (1.0,1.5) [txta] {a};
         \node (s1) at (1.5,1.5) [txta] {a};
         \node (s1) at (2.0,1.5) [txta] {a};
         \node (s1) at (2.5,0.7) [txtb] {b};
         \node (s1) at (3.0,0.7) [txtb] {b};
         \node (s1) at (3.5,0.7) [txtb] {b};
         \node (s1) at (4.0,1.5) [txta] {a};
        \end{tikzpicture}
        \caption{Dense coding}
     \end{subfigure}%
     \vspace{0.1cm}~
     \begin{subfigure}[t]{0.49\textwidth}
         \centering
        \begin{tikzpicture}[samplea/.style={inner sep=0.5mm,circle,fill=red},sampleb/.style={inner sep=0.5mm,circle,fill=blue},txta/.style={color=red},txtb/.style={color=blue}]
         \draw[-Latex] (0,0) to (5,0);
         \node at (5,-0.4) {Time};
        \node (s1) at (0.25,1.2) [samplea] {};
         \node (s4) at (2.25,1.2) [samplea] {};
         \node (s5) at (2.25,0.4) [sampleb] {};
         \node (s7) at (3.75,0.4) [sampleb] {};
         \node (s8) at (3.75,1.2) [samplea] {};
         \draw[txta] (s1) to (s4);
         \draw[txtb] (s5) to (s7);
         \node (fake) at (4.5,1.2) {};
         \draw[txta] (s8) to (fake);
         \node (s1) at (2.0,1.5) [txta] {a:$\Delta_t=4$};
         \node (s1) at (3.5,0.7) [txtb] {b:$\Delta_t=3$};
         \node (s1) at (4.7,1.5) [txta] {a:$\Delta_t=\dots$};
         \end{tikzpicture}
         \caption{Event-based coding}
     \end{subfigure}
     \caption{Dense and event-based coding of the same time-series. An event-based coding is more efficient than a dense coding at encoding sequences where the transmitted symbol changes only sparsely.}
     \label{fig:coding}
 \end{figure}
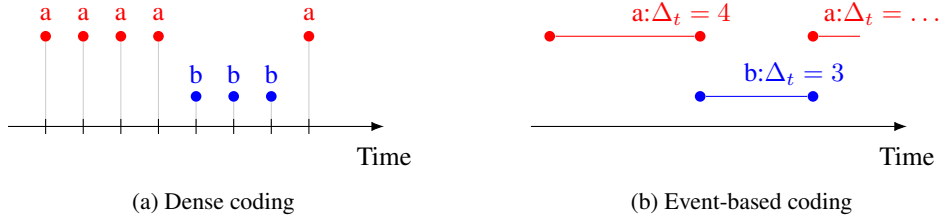

\textbf{Bit-stream XOR dataset}
Every data point is a block of 32 random bits. The binary labels are created by applying an XOR function on the bit block, i.e., class A if the number of 1s in the bit-stream are even, class B if the number of 1s in the bit-stream is odd. 
For training, a cross-entropy loss on these two classes is used. 
The training set consists of 100,000 samples, which are less than $0.0024\%$ of all possible bit-streams that can occur.
The test set consists of 10,000 samples. 

For the event-based encoding, we introduce a time-dimension.
The time is normalized such that the complete sequence equals 1 unit of time, i.e., 32 bits corresponds to exactly 1 second. 
An illustration of the two different encodings is shown in Figure \ref{fig:coding}.

\textbf{Person Activity} We consider a variation of the "Human activity" dataset described in \cite{rubanova2019latent} form the UCI machine learning repository \cite{dua2019}. 
The dataset is comprised of 25 recordings of human participants
performing different physical activities.
The eleven possible activities are ''lying down'', ''lying'', ''sitting down'', ''sitting'', ''standing up from lying'', ''standing up from'', ''sitting'', ''standing up from sitting on the ground'', ''walking'', ''falling'', ''on all fours'', and ''sitting on the ground''. 
The objective of this task is to recognize the activity from inertial sensors worn by the participant, i.e., a per-time-step classification problem.
We group the eleven activities listed above into seven different classes, as proposed by \cite{rubanova2019latent}. 

The input data consists of sensor readings from four inertial measurement units placed on the participant's arms and feet.
The sensors are read at a fixed period of 211 ms but have different phase-shifts in the 25 recordings. Therefore, we treat the data as irregularly sampled time-series.

The 25 recordings are split into partially overlapping sequences of length 32, to allow an efficient training of the machine learning models. 

Our results are not directly comparable to the experiments in \cite{rubanova2019latent}, as we use a different representation of the input features. While \cite{rubanova2019latent} represents each input feature as a value-mask pair, i.e., 24 input features, we represent the data in the form of a 7-dimensional feature vector.
The first four entries of the input indicate the senor ID, i.e., which arm or foot, whereas the remaining three entries contain the sensor reading.

\textbf{Event-based seqMNIST}
The MNIST dataset consists of 70,000 data points split into 60,000 training and 10,000 test samples \cite{lecun1998gradient}. 
Each sample is a 28-by-28 grayscale image, quantized with 8-bits and represents one out of 10 possible digits, i.e., a number from 0 to 10.

We pre-process each sample as follows:
We first apply a threshold to transform the 8-bits pixel values into binary values. The threshold is 128, on a scale where 0 represents the lowest possible and 255 the larges possible pixel value.
We further transform the 28-by-28 image into a time-series of length 784. 
Next, we encode binary time-series in a event-based format. Essentially, the encoding step gets rid of consecutive occurrences of the same binary value, i.e., $1,1,1,1$ is transformed into $(1,t=4)$.
By introducing a time dimension, we can compress the sequences from 784 to an average of 53 time-steps.

To allow an efficient batching and training, we pad each sequence to a length of 256. Note that no information was lost during this process.
We normalize the added time dimension such that 256 symbols correspond to 1 second or unit of time.
The resulting task is a per-sequence classification problem of irregularly sampled time-series.

\textbf{Walker2d kinematic modeling}
Here we create a dataset based on the \texttt{Walker2d-v2} OpenAI gym \cite{gym} environment and the MuJoCo physics engine \cite{todorov2012mujoco}.
Our objective is to benchmark how well the RNN architecture can model kinematic dynamical systems in an irregularly sampled fashion.
The learning setup is based on an auto-regressive supervised learning, i.e., the model predicts the next state of the Walker2d environment based on the current state.

In order to obtain interesting simulation rollouts, we trained a non-recurrent policy by Proximal Policy Optimization (PPO) \cite{schulman2017proximal} using the Rllib \cite{liang2018rllib} reinforcement learning framework. We then collect the training data for our benchmark by performing rollouts on the \texttt{Walker2d-v2} environment using our pre-trained policy. Note that because the policy is deterministic, there is no need to include the actions produced by the policy in the training data.

We introduce three sources of uncertainty to make this task more challenging.
First of all, for each rollout we uniformly sample a checkpoint of policy at 562, 822, 923, or 1104 PPO iterations.
Secondly, we overwrite 1\% of all actions by random actions.
Thirdly, we exclude 10\% of the time-steps, i.e., we simulate frame-skips/frame-drops. Note that the last step transforms the rollouts into irregularly sampled time-series and introduces a time dimension.

In total, we collected 400 rollouts, i.e., 300 used for training, 40 for validation, and 60 for testing. For an efficient training, we align the rollouts into sequences of length 64. We use the mean-square-error as training loss and evaluation metric. We train each RNN for 200 epochs and log the validation error after each training epochs. At the end, we restore the weights that achieved the best (lowest) validation error and evaluate them on the test set.

\begin{table}[]
    \centering
    \caption{ODE-solvers used for the different RNN architectures involving ordinary differential equations}
\begin{tabular}{ccc}
\toprule
Model & ODE-solver & Time-step ratio \\
\midrule
CT-RNN & 4-th order Runge-Kutta & 1/3 \\
ODE-RNN & 4-th order Runge-Kutta & 1/3 \\
GRU-ODE & Explicit Euler & 1/4 \\
ODE-LSTM & Explicit Euler & 1/4 \\
\bottomrule
\end{tabular}
    \label{tab:ode_solvers}
\end{table}

\begin{table}[]
    \centering
    \caption{Hyperparameters}
\begin{tabular}{lcl}
\toprule
Parameter & Value & Description \\
\midrule
RNN latent dimension & 64 & number of neurons in the RNN\\
Minibatch size & 256 & \\
Optimizer & RMSprop \cite{tieleman2012lecture} & \\
Learning rate & 5e-3 & \\
Training epochs & 500/200 & Synthetic/real datasets \\
\bottomrule
\end{tabular}
    \label{tab:hyperparams}
\end{table}

\textbf{Reproducibility statement}
We publish all code and data used in our experimental setup at this link \url{https://github.com/mlech26l/ode-lstms}.

\end{document}